\documentclass[preprint,5p,times,twocolumn,authoryear]{elsarticle}

\usepackage{amssymb}

\usepackage[normalem]{ulem}
\usepackage{xcolor}
\usepackage{url}
\usepackage{dirtytalk}
\usepackage{amsmath,amsfonts}
\usepackage{multirow}
\usepackage{multicol}
\usepackage{hhline}
\usepackage{booktabs}
\usepackage{adjustbox}

\begin{document}

\begin{frontmatter}



\title{Domain-informed graph neural networks:\\a quantum chemistry case study}


\author[LIS,Swansea_CS]{Jay Morgan\corref{cor2}}
\author[LIS,Swansea_CS]{Adeline Paiement\corref{cor1}}
\author[Rostock1,Rostock2,Swansea_chemistry]{Christian Klinke\corref{cor3}}

\cortext[cor1]{Corresponding author, adeline.paiement@univ-tln.fr}
\cortext[cor2]{jay.morgan@univ-tln.fr}
\cortext[cor3]{christian.klinke@uni-rostock.de}

\affiliation[LIS]{organization={Université de Toulon, Aix Marseille Univ, CNRS, LIS},
            city={Marseille},
            country={France}}

\affiliation[Swansea_CS]{organization={Department of Computer Science, Swansea University},
             city={Swansea},
             postcode={SA2 8PP},
             country={United Kingdom}}

\affiliation[Rostock1]{organization={Institute of Physics, University of Rostock},
             city={Rostock},
             postcode={18059},
             country={Germany}}

\affiliation[Rostock2]{organization={Department “Life, Light \& Matter”, University of Rostock},
             city={Rostock},
             postcode={18059},
             country={Germany}}

\affiliation[Swansea_chemistry]{organization={Department of Chemistry, Swansea University},
             city={Swansea},
             postcode={SA2 8PP},
             country={United Kingdom}}

\begin{abstract}
We explore different strategies to integrate prior domain knowledge into the design of a deep neural network (DNN). We focus on graph neural networks (GNN), with a use case of estimating the potential energy of chemical systems (molecules and crystals) represented as graphs. We integrate two elements of domain knowledge into the design of the GNN to constrain and regularise its learning, towards higher accuracy and generalisation. First, knowledge on the existence of different types of relations (chemical bonds) between atoms is used to modulate the interaction of nodes in the GNN. Second, knowledge of the relevance of some physical quantities is used to constrain the learnt features towards a higher physical relevance using a simple multi-task paradigm. We demonstrate the general applicability of our knowledge integrations by applying them to two architectures that rely on different mechanisms to propagate information between nodes and to update node states.
\end{abstract}




\end{frontmatter}


\section{Introduction}
\label{sec:introduction}


We investigate the introduction of domain knowledge into the design of a graph neural network (GNN), to constrain and regularise its learning towards higher accuracy and generalisation. GNNs were first proposed in \citep{Scarselli2009} to process data represented as graph. An internal state $\mathbf{h}_v$ is associated to each node $v$ of the graph, and the ensemble of internal states serves to produce an output $\hat{y}$. Each internal state is iteratively updated, based on 1) input features $\mathbf{x}_v$ associated to the node, 2) input features $\mathbf{x}^e_{vw}$ associated to the edges of the node, and 3) actions from neighbour nodes $w$, represented as a message $\mathbf{m}_v=\sum_w \mathbf{m}_{vw}$. The message $\mathbf{m}_{vw}$ is generated by a message function $M(\mathbf{h}_v, \mathbf{h}_w, \mathbf{x}^e_{vw})$\footnote{This is a general definition, and in some implementations $M$ may only use a subset of its inputs.} using the internal state $\mathbf{h}_w$ of the neighbouring node. The message function $M$ is shared by all nodes. It is typically implemented by a perceptron, e.g. as in \citep{Liu2021} where a perceptron is applied to a concatenation of $\mathbf{x}^e_{vw}$ and of $\mathbf{h}_v\circ\mathbf{h}_w$ ($\circ$ being element-wise multiplication).
The update function $U(\mathbf{h}_v, \mathbf{x}_v, \mathbf{x}^e_{vw}, \mathbf{m}_v)$\footnote{Here also this is a general definition, and in some implementations $U$ may only use a subset of its inputs.} is also shared by all nodes. It is originally implemented by a perceptron, but it can also benefit from a recurrent kernel such as GRU (e.g. in \citep{Gilmer2017}) or LSTM.
Depending on the task, the GNN's output is then computed from the node states by a readout function, which may be implemented by a perceptron or more complex kernels.
Variants of the original GNN have been proposed to accommodate special graph types, such as directed, see e.g. \citep{Wu2019,Zhou2019} for an overview.

\begin{figure*}
    \centering
    \includegraphics[width=\linewidth]{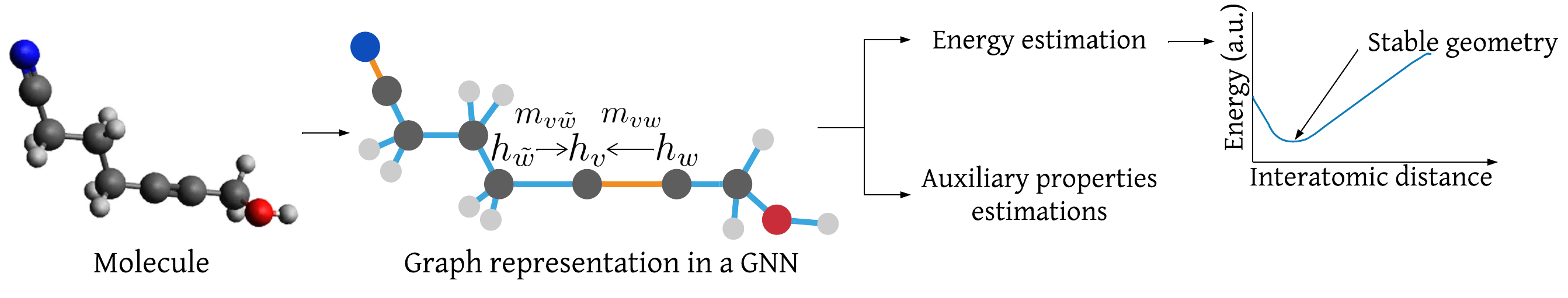}
    \caption{Graph representation of a molecule (left from \citep{Glavatskikh2019}) for GNN estimation of potential energy and stable geometry. Nodes of states $h_i$ are atoms and colour denotes atom type. Edges link chemically bonded atoms and colour denotes bond type. Non-bonded atoms may also share edges in fully connected graphs, but these edges are not represented for readability. Messages $m_{ij}$ are exchanged across edges. The GNN outputs an energy estimate, along with the estimates of one or several auxiliary (physical and chemical) properties in the case of multi-task learning. Energy estimates at different molecule geometries may be used to identify stable configurations at the minimum of energy.
    }
    \label{fig:overview}
\end{figure*}

\citet{Schlichtkrull2018} considered the case of edges representing different types of relation between nodes, and proposed to specialise the message function $M$ with regards to relation type $r$. This was achieved with distinct perceptrons. In \citep{Zhang2019}, Zhang et al. obtained a similar specialisation of $M$ to the relation type using specialised weights. In \citep{Chen2021}, Chen et al. took a different approach to account for relation type, with separate and specialised GNNs for each sub-graph of a given edge type, before final fusion of learnt representations.

We further explore avenues for integrating domain knowledge into the design of GNNs, with the aim to constrain and simplify the GNN's learning to improve its generalisation and accuracy, and to allow training on smaller datasets. We consider two distinct aspects, namely 1) the information flow within the GNN, and 2) the relevance of learnt node states for the application domain. Regarding point 1), we expand and generalise on the initial proposition of \citet{Schlichtkrull2018} through the exploration of different pathways for specialising the information flow within the GNN, namely by acting on the message generation $M$ or on the node update $U$. We propose a more general formulation that applies to all GNN architectures and their implementations of $M$ and $U$. With point 2), we also exploit the knowledge that some auxiliary quantities/qualities are closely related to the studied phenomena and should play a role in the internal representation of the DNN.

We consider the case study of estimating \emph{potential energies}\footnote{determined at the electronic ground-state at given positions of atoms (i.e. given geometry) for static systems} of a chemical system, either molecule or crystal, as a function of geometry (i.e. position of atoms). Such systems are represented as chemical graphs, with nodes denoting individual atoms, and edges the type of bond between them, as illustrated in Fig.~\ref{fig:overview}. For \emph{out-of-equilibrium (OoE)} systems, the atoms are at positions which are not at the minimum of potential energy. There is an interest in estimating the potential energy of OoE systems, as this allows searching for stable geometries through minimisation of the energy, with a possible aim to discovering new materials or simulating crystal growth. Such calculations typically require large amounts of resources and do not scale well to larger system sizes (i.e. number of atoms) \citep{Jiang2003,Erba2017}. This was addressed in the two seminal works MPNN \citep{Gilmer2017} and SchNet \citep{Schutt2017Quantum-chemicalNetworks} with two GNNs that estimate energy in a matter of milliseconds.
Indeed, GNNs and their message passing principle are well suited to capture the atomic interactions that underpin the system's chemical properties. Further recent works have built on these early GNNs, e.g. \citep{DeepMoleNet,PaiNN} improve on MPNN and \citep{PhysNet,DimeNet,DimeNet++,NequIP} improve on SchNet.

We summarise our main contributions as follows:

\textbf{[C1]} We leverage the existence of different edge types to modulate the information flow within the GNN (Section \ref{subsec:bond}). To this end, we formulate and compare two strategies, namely specialised message production and specialised node update. For the specialised node update, we provide different implementations for GRU and for dense layer-based update functions. While specialised message production was proposed in some previous works and for specific architectures, our formulation is more general, and our proposition of specialised node update is novel.

\textbf{[C2]} We leverage a multi-task learning framework to enforce learnt representations to be more related to quantities of interest (Section \ref{subsec:predictions}). We explore the potential of this approach to better capture the underlying mechanisms behind the studied phenomenon.

\textbf{[C3]} We evaluate our domain knowledge integration strategies both individually and jointly, on the case study of estimating the potential energy of chemical systems (molecules and crystals). 
Our methods are applied to the very different architectures of MPNN and SchNet to demonstrate their flexibility to GNN type (Section \ref{sec:results}). We demonstrate that they bring an improved accuracy and greater generalisation to graph structure and size. 
MPNN and SchNet are the basis for most recent models within our case study. Therefore, our improved performances for these base models suggest that our methods may be applied to these more recent models with similar improvements. While our knowledge integration strategies are demonstrated on potential energy estimation, they may be applied more generally to other application domains where graph representations are relevant, for example circulation of goods and people in economics \citep{Panford2020}, traffic prediction \citep{Diehl2019}, or E-commerce recommendations \citep{Liu2021}.

\textbf{[C4]} To support these experiments, we release three new datasets of OoE molecules and crystals of various complexities (Section \ref{subsec:datasets}).


\section{Previous works}
\subsection{GNNs for estimating potential energies of chemical systems}
\label{sub:base_models}

\citet{Gilmer2017} implement with their Message Passing Neural Network (MPNN)\footnote{We use the implementation from: \url{https://github.com/priba/nmp_qc} (MIT license).}
a rather classical GNN with atom type as node feature, interatomic distance as edge feature, a perceptron as message function, and GRU as update function. The internal state $\mathbf{h}_v$ is initialised as the node feature $\mathbf{x}_v$. Then, messages and node states at iteration $t+1$ are computed as:
\begin{equation}
\mathbf{m}_v^{t+1} = \sum_{w \in \mathcal{N}_v} M\left(\mathbf{h}_w^t, \mathbf{x}^e_{vw}\right)
\label{eq:message-function}
\end{equation}
\begin{equation}
\mathbf{h}_v^{t+1} = U\left(\mathbf{h}_v^t, \mathbf{m}_v^{t+1}\right)
\label{eq:update-function}
\end{equation}
with $\mathcal{N}_v$ the set of neighbours of atom $v$. Three iterations are performed, then the set of node states at all timesteps is used to estimate the system's potential energy as the sum of individual nodes' contributions, computed by perceptrons, using a mean-squared error loss.
In basic MPNN, only bonded atoms exchange messages ($\mathcal{N}_v$ only contains bonded neighbours). In a fully connected graph variant, which we denote as Extensive MPNN (E-MPNN), all pairs of atoms exchange messages to account for long-range interactions.
\citet{Gilmer2017} also introduced bond type (BT) information as edge input feature. This additional feature improved energy estimates over using interatomic distance alone. We refer to this other variant as MPNN-BTF (MPNN with Bond Type Feature).

With SchNet\footnote{\url{https://github.com/atomistic-machine-learning/schnetpack} (MIT license)}, \citet{Schutt2017Schnet:Interactions}
implement another GNN with 3 layers of interactions that do not share weights, as opposed to the recurrent iterations of MPNN, but weights are still shared between nodes/atoms for each layer. The interaction layers are followed by atom-wise dense layers, non-linearities, and a final sum pooling to obtain the energy estimate. Each interaction layer $l$ considers the sum of actions of neighbours on atom $v$ in a fully connected graph (Eq.~\ref{eq:message-function-schnet}).
The atoms' actions (i.e. messages) are element-wise multiplication $\circ$ of their node states $\mathbf{h}_w^l$ (updated by a dense layer with parameters $\mathbf{W}^l$, $\mathbf{b}^l$) with a radial filter $R^l$ that depends on the distance $d_{vw}$ between the two atoms. $R^l$ is implemented by two softplus dense layers. Node states are initialised based on atom type (i.e. node feature), then updated by a dense layer-based update function (Eq.~\ref{eq:update-function-schnet}).
\begin{equation}
\mathbf{m}_v^{l+1} = \sum_{w \in \mathcal{N}_v} \left(\mathbf{W}^l \mathbf{h}_w^l + \mathbf{b}^l\right) \circ R^l\left( d_{vw}\right)
\label{eq:message-function-schnet}
\end{equation}
\begin{equation}
\mathbf{h}_v^{l+1} = \mathbf{h}_{v}^{l} + V^l\left(\mathbf{h}_v^l, \mathbf{m}_v^{l+1}\right)
\label{eq:update-function-schnet}
\end{equation}

\subsection{Domain-informed design of a deep neural network}

In \citep{Gilmer2017,Schutt2017Schnet:Interactions,DeepMoleNet,PaiNN,PhysNet,DimeNet,DimeNet++,NequIP}, the choice of a GNN architecture is a well-motivated inductive bias where the internal representation and information flow are designed to fit with the physics of the replicated phenomenon, namely the interactions of atoms producing the system's potential energy. Other similar examples of domain-informed choice of a GNN architecture include \citep{mrowca2018flexible} that predicts future states of deformable objects represented by a hierarchical graph that decomposes them into particles at various scales. Convolution operations are defined on this graph to apply external forces to the system, and to exchange information on collisions and physical state change. Similarly, \citep{Kawahara2017} defined convolution operations on a brain connectivity graph to predict neurodevelopment scores. \citet{Diehl2019} found that accounting for interactions between neighbouring vehicles helps predict short-term behaviours of traffic participants. When predicting bilateral trade, \citet{Panford2020} hypothesised that \say{adoption of specific domestic trade policies in one country can influence similar policies to its neighbours}, which was realised as a graph to represent trade relationships between countries. \citet{Liu2021} also used a GNN architecture for E-commerce recommendation where missing relationships are inferred from a partially connected graph.

Other works have adapted the training of deep neural networks (DNN) based upon prior physics knowledge. \citet{Raissi2018} constrained a dense-layer DNN that estimates physical quantities (e.g. velocity, pressure) by a loss term obtained from partial differential equations of fluid dynamics.
These equations are implemented using automatic differentiation, and their parameters are learnable.
\citet{Schutt2017Schnet:Interactions} used a similar approach to improve SchNet's predictions of both energies and their derivatives w.r.t. atom positions, using a loss term based on computed interaction forces. In the present work, we do not employ equations of atomic interactions, as they become intractable for larger systems.
Instead, we focus on the domain-informed architecture avenue, i.e. how a GNN's design may further account for the known properties (e.g. known physics in our case study) of the problem.

\section{Proposed domain knowledge integration}
\label{sec:methodology}

\subsection{Specialising interactions based on edge type}
\label{subsec:bond}

The types of relation between nodes of a graph are an important factor when considering the nodes' interactions towards the GNN's inference. As an illustration, in our case study, the types of bonds between atoms determine atomic interactions and their contribution to the system's energy. Therefore, when estimating the potential energy of a system as a function of geometry, it may be beneficial to account for the contribution of different bond types (BT). MPNN-BTF provide a clue towards confirming this assertion. By introducing BT information as edge input feature in their GNN, \citet{Gilmer2017} improved energy estimates over using distance alone. We take this domain knowledge integration principle further, and we propose to design the information exchange within the GNN to reflect the relations between nodes. In our case study, this results in the nodes' information exchange better reflecting the physics of atomic interactions.

Similar to \citep{Schlichtkrull2018}, we introduce \emph{specialised interactions based on relation type}, with relation type being BT\footnote{BT is predetermined using RDKit, \url{https://www.rdkit.org/} (BSD 3-Clause license), and Antechamber \citep{Wang2001Antechamber:Calculations} (GPL-3 license), for molecules with Canonical SMILES representation and for others, respectively. BT is provided as an input and it is not determined by the GNN.} in our case study. While \citep{Schlichtkrull2018} focused on the production of messages that are specialised to the relation, we explore two strategies of specialising either the message production, or the node update. These two processes operate based on their own and separate unit types (e.g. perceptron, GRU...) and learnt parameters. Thus, our two strategies act on different learning elements within the GNN. Furthermore, while \citep{Schlichtkrull2018} focused on a single GNN type (GCN), we provide more general formulations that are adapted to different architectures.

\subsubsection{Specialised message production}

Separate messages $\mathbf{m}_{vw}^{r}$ for each relation type $r$ between nodes are produced by a relation-specialised message function $M_r$. In the specific case of a perceptron-based $M_{r}$, this is equivalent to \citep{Schlichtkrull2018}. For other implementations of the message function, the $r$-specialised version is simply obtained by having a separate set of learnt parameters per relation type $r$. For our two example GNNs, the $r$-specialised message function is denoted $M_{r}$ (MPNN, see Eq.~\ref{eq:message-function}) or $R^l_{r}$ (SchNet, see Eq.~\ref{eq:message-function-schnet}) and become respectively:
\begin{equation}
\mathbf{m}_v^{t+1} = \alpha \sum_{w \in \mathcal{N}_v} M\left(\mathbf{h}_w^t, \mathbf{x}^e_{vw}\right) + (1 - \alpha) \sum_{r \in \mathcal{R}} \sum_{w \in \mathcal{N}^{r}_v} M_{r}\left(\mathbf{h}_w^t, \mathbf{x}^e_{vw}\right)
\label{eq:augm-message-function}
\end{equation}
\begin{equation}
\mathbf{m}_v^{l+1} = \alpha \sum_{w \in \mathcal{N}_v} \mathbf{h}_w^l \circ R^l\left(d_{vw}\right) + (1 - \alpha) \sum_{r \in \mathcal{R}} \sum_{w \in \mathcal{N}^{r}_v} \mathbf{h}_w^l \circ R_{r}^l\left(d_{vw}\right)
\label{eq:augm-message-function-schnet}
\end{equation}
with $\alpha$ modulating the strength of the relation-specialised message production with regards to the original generic message production. $\mathcal{R}$ is the set of relation types. In our application scenario, it is the set of bond types, that may include a `no bond' element to consider a fully connected graph as in SchNet. $\mathcal{N}^{r}_v$ is the set of neighbour nodes that share a relation of type $r$ with node $v$.

\subsubsection{Specialised node update}

Specialised interaction may also use generic messages $\mathbf{m}_{vw}$, but handling them in a relation-specialised manner when updating node states. We explore the following different implementations:

\paragraph{Specialised weighting of the messages}
Messages coming from nodes of different relation types are weighted by a relation-specialised and learnable (scalar or vector) weight $\lambda_{r}$:
\begin{equation}
    \mathbf{m}_v = \alpha \sum_{w \in \mathcal{N}_v} \mathbf{m}_{vw} + (1 - \alpha) \sum_{r \in \mathcal{R}} \lambda_{r} \sum_{w \in \mathcal{N}_v^{r}} \mathbf{m}_{vw}
    \label{eq:weighted_sum_messages}
\end{equation}
In the specific case of a perceptron-based message function $M$, this approach is equivalent to the basis-decomposition regularisation proposed in \citep{Schlichtkrull2018} as a mean to reduce the number of learnable parameters associated to specialised message production.

\paragraph{Specialised update functions}
This approach implements separate relation-specialised update functions and sums their contributions. As for $M_r$, $r$-specialised update functions of any type may be obtained through specialised sets of learnt parameters. Node update for MPNN and SchNet become respectively:
\begin{equation}
    \mathbf{h}_v^{t+1} = \alpha U\left(\mathbf{h}_v^t, \mathbf{m}_v^{t+1}\right) + (1 - \alpha) \sum_{r \in \mathcal{R}} U_{r}\left(\mathbf{h}_v^t, \sum_{w \in \mathcal{N}_v^{r}} \mathbf{m}_{vw}^{t+1}\right)
    \label{eq:specialised_update}
\end{equation}
\begin{equation}
    \mathbf{h}_v^{l+1} = \mathbf{h}_v^l + \alpha V^l\left(\mathbf{h}_v^l, \mathbf{m}_v^{l+1}\right) + (1 - \alpha) \sum_{r \in \mathcal{R}} V^l_{r}\left(\mathbf{h}_v^l, \sum_{w \in \mathcal{N}_v^{r}} \mathbf{m}_{vw}^{l+1}\right)
    \label{eq:specialised_update_schnet}
\end{equation}

In the case of MPNN (Eq.~\ref{eq:specialised_update}), where the update function is implemented by a GRU unit, we experiment with three different levels of weight sharing when specialising $U_{r}$ in order to limit the amount of added model complexity. A similar strategy may be employed for other types of update function.

\subparagraph{Implementation 1)}
We use separate GRU cells, one per relation type, to implement the different $U_{r}$.

\subparagraph{Implementation 2)}
We use a single GRU cell, so that all relation channels share the GRU's internal state. The GRU's $\mathbf{W}_z$, $\mathbf{W}_r$ and $\mathbf{W}_h$ weight matrices contain weights that are specialised for the different relation types. In practice, this amounts to concatenating the different messages from each relation type 
$\begin{bmatrix} \sum_{w \in \mathcal{N}_v^1} \mathbf{m}_{vw} \hdots \sum_{w \in \mathcal{N}_v^{\left|\mathcal{R}\right|}} \mathbf{m}_{vw} \end{bmatrix}^T$
before providing them to the GRU cell.

\subparagraph{Implementation 3)}
We also use a single GRU cell with concatenated messages from different relation types, but we reduce the number of free GRU parameters in $\mathbf{W}_z$, $\mathbf{W}_r$ and $\mathbf{W}_h$ by sharing them between relation channels: 
$\mathbf{W}_{z/r/h} = \begin{bmatrix} \mathbf{Q}_{z/r/h} \hdots \mathbf{Q}_{z/r/h} \end{bmatrix}^T$
with $\mathbf{Q}_z$, $\mathbf{Q}_r$ and $\mathbf{Q}_h$ being matrices.

\subsection{Relating learnt features to relevant physical quantities}
\label{subsec:predictions}

Multi-task learning (MTL) may be seen as a way to introduce an inductive bias, where one or several auxiliary tasks further constrain the model and its learnt features (see e.g. \citep{Ruder2017,Zhang2018} for a review, and Fig.~\ref{fig:overview} for illustration). We use this paradigm to encourage the GNN's node states to relate more to relevant (physical) properties for our problem.
Drawing motivation from the fact that BT is characterised by the valence property of atoms, we hypothesise that a more physically relevant node state may better support BT-specialised interactions. We explore the potential of MTL for better relating learnt features to physical parameters that are relevant to a final task, and the effect on this task. Our exploration encompasses different GNN architectures and auxiliary tasks. We also evaluate separately the effects of individual tasks.

We experiment with the auxiliary estimation of low- and high-level system-wise properties:
1) the number of atoms of each type present in the system. This composition is directly relevant to the value of potential energy, therefore it may be beneficial to draw the attention of the DNN on composition.
2) the number of orbitals associated with each atom type. In addition to being physically relevant to the definition of potential energy, this is also particularly relevant to determine the BT between two atoms, in support for specialised interaction.
3) a probability distribution for the scaling to the stable geometry, estimated as a Gaussian function as in \citep{Timoshenko2018}. This quantity is less directly related to the physical properties of the chemical system, while still being related to potential energy. A chemical system at equilibrium being at its minimum of potential energy, the task of estimating how far the system is to equilibrium (or to minimum potential energy) is an interesting task that considers the whole chemical system and how it relates to its potential energy.
In future works, we may investigate auxiliary quantities linked to the physics of individual nodes/atoms, such as atomic forces or the medium-level descriptor atom-centred symmetry functions (ACSFs) used in \citep{DeepMoleNet}. In other case studies, a purpose-designed set of relevant properties would need to be designed, using knowledge of the problem. While we focus on properties that are relevant to the whole graph, node-wise properties may also be considered in the future.

In practice, for each auxiliary estimation, a new output layer is added and a mean-squared error loss term is minimised during training. This is identical to the energy output estimation, and the type of output layer depends on the GNN architecture (see Section \ref{sub:base_models} for details on the two GNNs used in our experiments). Each auxiliary loss term is weighted by an empirical $\beta = 0.3$ hyper-parameter while the original potential energy loss keeps a weight of $1$ to emphasis the main task more:
\begin{equation}
    L_{total}=L_{MSE}(y, \hat{y}) + \beta \sum_{a \in \mathcal{A}} L_{MSE}(a, \hat{a})
    \label{eq:loss_auxiliary}
\end{equation}
with $y$ and $\hat{y}$ the ground-truth and predicted potential energies, and $\mathcal{A}$ the set of auxiliary quantities $a$ being used.

\section{Experiments}
\label{sec:results}

We evaluate the impact of our knowledge integration on the two GNNs of MPNN and SchNet. They are the two bases to current state-of-the-art (SoTA) models for estimating potential energies e.g. \citep{DeepMoleNet,PaiNN,PhysNet,DimeNet,DimeNet++,NequIP}, thus our methods may also be applied to current and future SoTA, likely with similar results. In addition, the fact that these two architectures are very different suggests that similar improvements may be obtained on other GNN types too.

Neither models are used in the same conditions as in their original papers.
In \citep{Gilmer2017}, MPNN was only evaluated on stable configurations of molecules from the QM9 dataset \citep{blum,Montavon2013}, that contains a diverse set of 134k molecules using 5 atom types. We further evaluate it on new datasets of OoE molecules and crystals (Section \ref{subsec:datasets}).
Similarly, for higher accuracy, SchNet was trained in \citep{Schutt2017Schnet:Interactions} either on single molecules at a time (from the MD17 dataset \citep{Chmiela2017,Schutt2017Quantum-chemicalNetworks,Chmiela2018TowardsFields} of 8 small organic molecules with independent perturbations of their atoms' positions) or on isomers (i.e. molecules of same size and atomic composition, from the ISO17 dataset \citep{Schutt2017Schnet:Interactions,Schutt2017Quantum-chemicalNetworks,Ramakrishnan2014} of 129 isomers of C7O2H10). We apply SchNet to our harder scenario of jointly modelling multiple molecules and crystals of various sizes and atomic compositions. We also apply both GNNs to unseen sizes and compositions.

A comparison of MPNN and E-MPNN is provided in the sup. materials, where MPNN does as well as E-MPNN on one dataset, and outperforms it on all others. Thus, we only work with MPNN. We do not use MPNN-BTF because we aim to demonstrate the effectiveness of accounting for physics knowledge in the design of the GNN, rather than merely in the input features. Results of the augmented MPNN-BTF are still provided in the sup. materials.
We use the base models' default hyper-parameters for both original and augmented versions. For (base and augmented) MPNN, we experimented with different numbers of iterations (see sup. materials). MPNN's default of 3 worked best. New hyper-parameters $\alpha$ are learnable and optimised during training of the DNN. The sup. materials provide a list of the number of additional parameters brought by our methods. Early stopping at 50 stable epochs ensures that each DNN trains for a suitable time. The datasets are split between training, testing, and validation in proportions appropriate for the data complexity as detailed in the sup. materials.

\subsection{Datasets}
\label{subsec:datasets}

\subparagraph{Extended QM9 (E-QM9)} includes diverse sizes (i.e. number of atoms) and compositions of OoE molecules, through extending a subset of QM9 with OoE versions of 10k of its molecules.

\begin{figure}
    \centering
    \includegraphics[width=0.32\linewidth]{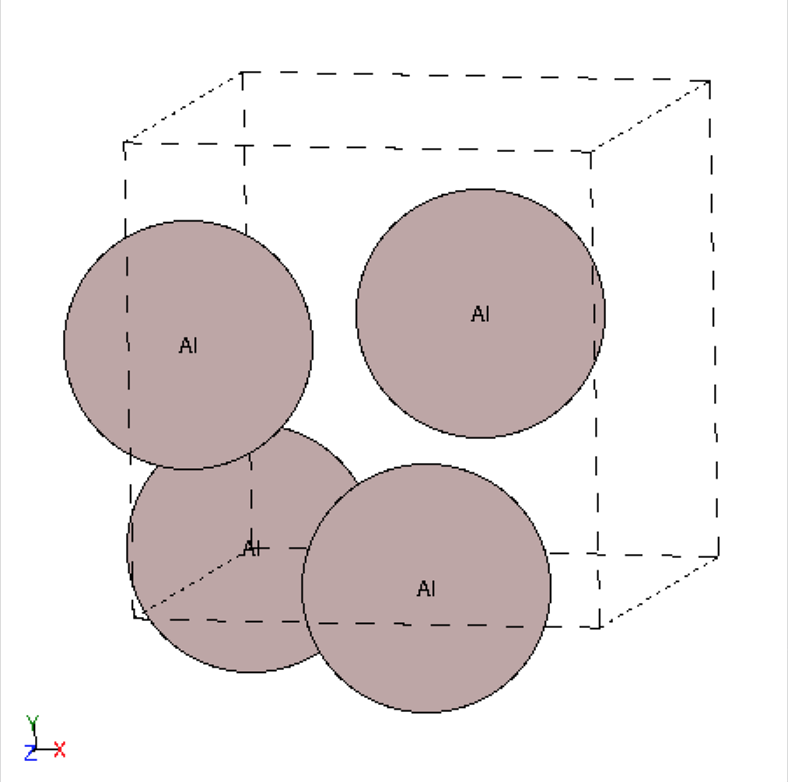}
    \includegraphics[width=0.32\linewidth]{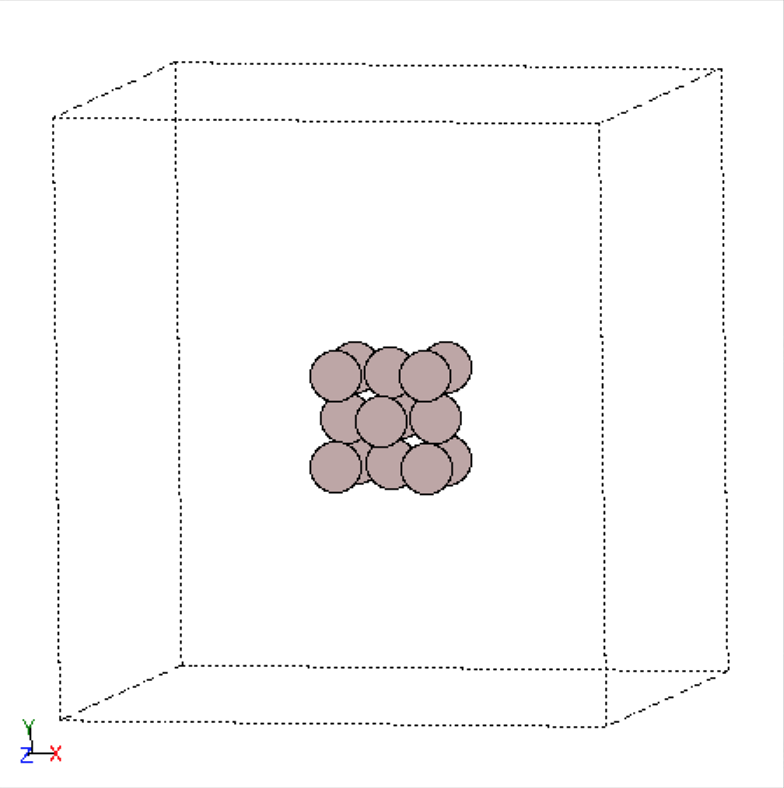}
    \includegraphics[width=0.32\linewidth]{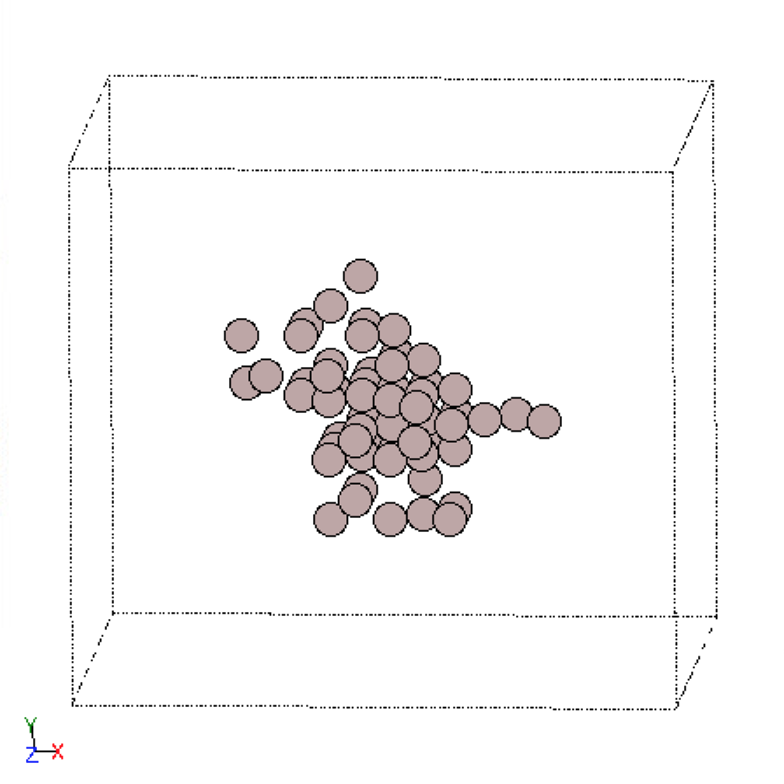}
    \caption{FCC Bravais crystal lattice (left) and growing  crystal structures (centre: seed, right: intermediate).}
    \label{fig:crystals}
\end{figure}

\subparagraph{Periodic crystals (PC)} allows learning regular bonding patterns that arise in periodic structures by repeating the base crystal lattice. We use the Face-Centred Cubic (fcc) Bravais lattice (Fig.~\ref{fig:crystals} left) for aluminium (Al) and copper (Cu) crystals.

\subparagraph{Crystal Growth (CG)} contains growing crystals of increasing size and complexity. Starting from a basic fcc crystal seed of 14 atoms (Fig.~\ref{fig:crystals} centre), new systems are generated by iteratively placing atoms at a random location on the surface of the growing crystal following its lattice pattern (Fig.~\ref{fig:crystals} right), with sizes ranging from 15 to 114 atoms. We use 20 random seeds for each atom type, thus creating 40 varied Al and Cu crystal growths and 4,000 stable systems. As a result, for a given crystal size and composition (atom type), there are 20 samples with differently located atoms. CG enables experimenting with large scale atomic interactions in non-regular systems, and enables evaluation of an ML method's ability to learn how each atom contributes to the final potential energy.

In all datasets, OoE systems are obtained by compressing/dilating all interatomic distances (i.e. isometrically) at regular intervals within 90-150\% of stable geometry, which we refer to as \emph{`scaling'}. In other words, scaling is applied to the coordinates of all atoms within the system: $\hat{\bf{x}}=\lambda\bf{x}$. At each geometry, the ground-truth potential energy is calculated using CP2K\footnote{\url{https://www.cp2k.org/} (under the GPL-2.0 license)}'s DFT.

\begin{table*}[h]
\centering
\caption{Impact of each domain knowledge integration strategy on MPNN (top) and SchNet (bottom) for E-QM9. Results are in the format: mean (std). The specialised interaction methods that optimise at best energy and geometry are highlighted in bold for each GNN. We also show in the last column the added number of parameters in percentage of original model size.}
\label{tab:branches}
\begin{tabular}{@{}lllrccc@{}c}
\toprule
 & \multicolumn{3}{c}{{{Strategy}}} & AE & RE & DSG & +\% parameters \\ \midrule
\tiny{1} & \multicolumn{3}{r}{\textbf{MPNN base model} with no BT information}                & 0.242 (1.318) & 0.0029 (0.015) & 0.034 (0.074) & -- \\
\tiny{2} & \multicolumn{3}{r}{MPNN-BTF}               & 0.091 (0.476) & 0.0012 (0.005) & 0.039 (0.056) & -- \\

\hhline{----~~~}
\tiny{3} & \multirow{6}{1.5cm}{$r$-specialised interactions} 
& \multicolumn{2}{r}{Specialised message Eq.~\ref{eq:augm-message-function} ($\alpha=0.624$)} & 0.272 (1.293) & 0.0034 (0.014) & 0.051 (0.080) & 398.20 \\
\tiny{4} & & \multirow{5}{1.5cm}{Specialised node updates} & Eq.~\ref{eq:weighted_sum_messages} scalar $\lambda_r$ ($\alpha=0.628$) & 0.122 (0.431) & 0.0016 (0.005) & 0.032 (0.057) & 5.28$e^{-4}$\\
\tiny{5} & & & Eq.~\ref{eq:weighted_sum_messages} vector $\lambda_r$ ($\alpha=0.627$) & 0.105 (0.502) & 0.0013 (0.005) & 0.050 (0.049) & 3.85$e^{-2}$\\
\tiny{6} & & & Eq.~\ref{eq:specialised_update} impl. 1) ($\alpha=0.615$) & \bf{0.106} \bf{(0.253)} & \bf{0.0014} \bf{(0.003)} & \bf{0.023} \bf{(0.048)} & 17.11\\
\tiny{7} & & & Eq.~\ref{eq:specialised_update} impl. 2) ($\alpha=0.635$) & \bf{0.073} \bf{(0.170)} & \bf{0.0010} \bf{(0.002)} & \bf{0.030} \bf{(0.048)} & 10.18 \\
\tiny{8} & & & Eq.~\ref{eq:specialised_update} impl. 3) ($\alpha=0.618$) & 0.131 (0.436) & 0.0017 (0.005) & 0.025 (0.055) & 3.42 \\

\hhline{----~~~}
\tiny{9} & \multirow{3}{1.5cm}{Auxiliary estimates of} & \multicolumn{2}{r}{\# atoms of each type}    & 0.171 (0.986) & 0.0021 (0.011) & 0.031 (0.046) & 3.43 \\
\tiny{10} & & \multicolumn{2}{r}{\# orbitals}            & 0.195 (0.545) & 0.0025 (0.006) & 0.033 (0.046) & 3.43 \\ 
\tiny{11} & & \multicolumn{2}{r}{distance to stable geometry}      & 0.119 (0.698) & 0.0015 (0.008) & 0.033 (0.046) & 3.40 \\
\midrule
\tiny{12} & \multicolumn{3}{r}{\textbf{SchNet base model}}    & 0.038 (0.037) & 0.0005 (0.0005) & 0.020 (0.031) & -- \\
\hhline{----~~~}
\tiny{13} & \multirow{4}{1.5cm}{$r$-specialised interactions} 
 & \multicolumn{2}{r}{Specialised message Eq.~\ref{eq:augm-message-function-schnet} ($\alpha=0.734$)} & 0.036 (0.035) & 0.0005 (0.0005) & 0.022 (0.032) & 220.43 \\
 
\tiny{14} & & \multirow{3}{1.5cm}{Specialised node updates} & Eq.~\ref{eq:weighted_sum_messages} scalar $\lambda_r$ ($\alpha=0.529$)  & \bf{0.020} \bf{(0.018)} & \bf{0.0003} \bf{(0.0002)} & \bf{0.025} \bf{(0.032)} & 8.01$e^{-3}$ \\

\tiny{15} & & & Eq.~\ref{eq:weighted_sum_messages} vector $\lambda_r$ ($\alpha=0.570$) & 0.024 (0.028) & 0.0003 (0.0003) & 0.034 (0.034) & 0.51 \\
 
\tiny{16} & & & Eq.~\ref{eq:specialised_update_schnet} ($\alpha=0.727$) & 0.031 (0.032) & 0.0004 (0.0004) & 0.015 (0.028) & 220.43 \\

\hhline{----~~~}
\tiny{17} & \multirow{3}{1.5cm}{Auxiliary estimates of} & \multicolumn{2}{r}{\# atoms of each type}         & 0.038 (0.036) & 0.0005 (0.0005) & 0.032 (0.033) & 0.39 \\
\tiny{18} & & \multicolumn{2}{r}{\# orbitals}    & 0.036 (0.035) & 0.0005 (0.0005) & 0.022 (0.032) & 0.39 \\ 
\tiny{19} & & \multicolumn{2}{r}{distance to stable geometry}    & 0.049 (0.044) & 0.0007 (0.0006) & 0.037 (0.033) & \ 0.36 \\
\toprule
\end{tabular}
\end{table*}

We generate two subsets of CG: \emph{Stable Crystal Growth} (SCG) with only stable geometries, and \emph{Unstable Crystal Growth} (UCG) with scaling. Both contain Al and Cu crystals. SCG aims to evaluate an ML model at varying system sizes (i.e. number of atoms) without the added complexity of varying scale. For UCG, we select 5 of the random crystal growth seeds for each atom type to consider 1,000 stable geometries to be scaled. Statistics on all datasets are in the sup. materials. The datasets are publicly available at \url{https://doi.org/10.6084/m9.figshare.12360620}.

\subsection{Evaluations of the individual domain knowledge integration strategies}
\label{subsec:physics_integration_test}

We evaluate the individual effects of the different knowledge integration methods in comparison to non-augmented base GNNs, on E-QM9.
Table~\ref{tab:branches} reports absolute error (AE) and relative error ($\text{RE}=\frac{|\hat{y} - y|}{|y|}$) between true $y$ and predicted $\hat{y}$ energies in a.u. unit. Squared errors are provided in the sup. materials. We also report the Distance to Stable Geometry (DSG) which is the absolute difference between the scaling where the predicted energy is minimal (optimised by multiple DNN queries for various scalings), and the ground-truth at-equilibrium scaling (i.e. 100\%): $\text{DSG}=|\lambda-1|$.
Mean and std are computed over chemical systems, with std indicating the ability to handle a large variety of systems.

Results of (non-augmented) base models are provided in rows 1,12 of Table~\ref{tab:branches}.
When compared against these base models, and considering only the best performing methods for $r$-specialised interactions, all integrations of domain knowledge tend to improve energy estimation and/or finding stable geometries. Since all the base and augmented models are trained on the exact same data (no data augmentation), these improvements over base models cannot come from training on an extension of QM9 that contains OoE molecules. Instead, improvements reflect the effect of integrating domain knowledge into the design of the augmented GNNs.

The impact of BT information is the strongest for both architecture types (rows 2,6,7,14). This confirms \citet{Gilmer2017}'s observation that BT is relevant for estimating energy. Furthermore, accounting for BT in the design of $r$-specialised interactions in MPNN (rows 6,7) has a stronger positive impact than merely using it as an input edge feature as in \citep{Gilmer2017} (row 2). This suggests that $r$-specialised interactions better capture the physics of atomic interactions.

When examining the effects of the different $r$-specialised interaction methods in parallel to the additional model complexity that they bring (last column of Table~\ref{tab:branches}, see also sup. materials for more numbers of parameters), we notice that the very simple $r$-specialised weighting of the messages (Eq.~\ref{eq:weighted_sum_messages}) performs generally well while bringing a very limited number of extra parameters. On the other hand, $r$-specialised message production (Eqs.~\ref{eq:augm-message-function} and \ref{eq:augm-message-function-schnet}) brings a very large number of new parameters (+220\% and +398\%), which may explain its lower performance. The same observation applies to $r$-specialised update functions for SchNet (Eq.~\ref{eq:specialised_update_schnet}, +220\%). However, in the case of MPNN, $r$-specialised update functions (Eq.~\ref{eq:specialised_update}) provide a lower number of new parameters (+3\% to 17\%), and remain within a manageable complexity that is exploited to bring the best performance improvement. In fact, the two implementations that bring the highest increase in complexity (+10-17\%) provide the best performance. Considering that MPNN and SchNet were originally proposed with different sizes of node states (73 vs. 128), it is also a possible explanation for MPNN benefiting more than SchNet from an added model complexity.

\begin{table*}[h]
\centering
\caption{Ablation study and effect on handling different atom types in crystals of UCG}
\label{tab:ablation}
\begin{tabular}{rcccc}
\toprule
\multicolumn{1}{c}{ } & \multicolumn{2}{c}{Al} & \multicolumn{2}{c}{Cu} \\
\cmidrule(l{3pt}r{3pt}){2-3} \cmidrule(l{3pt}r{3pt}){4-5}
Method & AE & DSG & AE & DSG\\
\midrule
MPNN & 0.643 (0.495) & 0.096 (0.135) & 6.250 (4.772) & 0.214 (0.192) \\
Augm.-MPNN & 0.150 (0.120) & 0.031 (0.037) & 1.160 (0.878) & 0.108 (0.122)\\
Augm.-MPNN w/o $r$-spec. interactions & 0.180 (0.144) & 0.050 (0.036) & 0.638 (0.582) & 0.071 (0.071)\\
Augm.-MPNN w/o aux. \# atoms & 0.179 (0.129) & 0.011 (0.026) & 0.928 (0.760) & 0.098 (0.125)\\
Augm.-MPNN w/o aux. \# orbitals & 0.190 (0.141) & 0.020 (0.034) & 0.906 (0.724) & 0.080 (0.073)\\
Augm.-MPNN w/o aux. DSG & 0.177 (0.151) & 0.035 (0.038) & 0.834 (0.680) & 0.082 (0.102)\\
\midrule
SchNet & 6.357 (3.159) & 0.117 (0.047) & 5.355 (5.278) & 0.066 (0.088)\\
Augm.-SchNet & 0.333 (0.247) & 0.069 (0.071) & 0.368 (0.259) & 0.002 (0.008)\\
Augm.-SchNet w/o $r$-spec. interactions & 0.444 (0.358) & 0.080 (0.073) & 1.424 (1.490) & 0.034 (0.060)\\
Augm.-SchNet w/o aux. \# atoms & 0.395 (0.169) & 0.102 (0.025) & 0.304 (0.238) & 0.001 (0.009)\\
Augm.-SchNet w/o aux. \# orbitals & 0.241 (0.146) & 0.006 (0.019) & 0.286 (0.176) & 0.054 (0.031)\\
Augm.-SchNet w/o aux. DSG & 0.296 (0.203) & 0.035 (0.049) & 0.328 (0.220) & 0.000 (0.000)\\
\bottomrule
\end{tabular}
\end{table*}

\begin{table*}[h]
    \centering
    \caption{Evaluation of the base (left) and augmented (right) GNNs. For each architecture type, best results between base and augmented GNNs are highlighted in bold.}
    \label{tab:datasets}
    \begin{tabular}{cccccc}
    \toprule
         & & \multicolumn{2}{c}{Base} & \multicolumn{2}{c}{Augmented} \\
         \cmidrule(lr){3-4} \cmidrule(lr){5-6}
        Model & Dataset & AE & DSG & AE & DSG \\
    \midrule
    \multirow{4}{*}{MPNN} & E-QM9 & 0.242 (1.318) & 0.034 (0.074) & \textbf{0.065 (0.161)} & \textbf{0.030 (0.045)} \\
    & Periodic Crystals                & 0.041 (0.032) & \textbf{0.073 (0.079)} & \textbf{0.039 (0.035)} & 0.074 (0.079) \\
    & Stable CG                   & 2.796 (3.797) & --            & \textbf{2.105 (2.624)} & --            \\
    & Unstable CG                 & 2.892 (3.820) & 0.344 (0.083) & \textbf{0.655 (0.805)} & \textbf{0.069 (0.098)} \\
    \midrule
    \multirow{4}{*}{SchNet} & E-QM9 & 0.038 (0.037)   & 0.020 (0.031) & \textbf{0.026 (0.025)} & \textbf{0.016 (0.028)} \\
    & Periodic Crystals                      & \textbf{13.444 (15.984)} & 0.054 (0.012) & 15.961 (19.424) & \textbf{0.039 (0.053)} \\
    & Stable CG                         & 2.021 (1.829)   & --            & \textbf{1.217 (1.586)} & -- \\
    & Unstable CG                       & 5.866 (4.377)   & 0.091 (0.075) & \textbf{0.351 (0.254)} & \textbf{0.035 (0.060)} \\
    \toprule
    \end{tabular}
\end{table*}

The auxiliary estimation of physical properties (rows 9-11) also improves on base MPNN, but less so than BT information. On SchNet, estimating the number of orbitals also provides a slight improvement (row 18).
These improvements may come from the models implicitly discovering and encoding useful physics representations required for accurate predictions.
Indeed, as discussed in Section \ref{subsec:predictions}, these quantities are chosen based on knowledge of the domain to help focus the GNN's attention on important aspects of the learning problem. In particular, the distance to stable geometry is related to a high level system optimisation and its improved performance may indicate that the GNN's features better capture the dependency of energy on geometry.
In future works, we may investigate auxiliary quantities linked to the physics of individual atoms/nodes.

We further explore the effect of the best method for $r$-specialised interactions (Eq.~\ref{eq:specialised_update} impl. 2 for MPNN and Eq.~\ref{eq:weighted_sum_messages} with scalar $\lambda_r$ for SchNet) and of each auxiliary estimation in an ablation study using UCG (results are reported in Table \ref{tab:ablation}).
As previously, the two GNN architectures don't benefit equally from a same augmentation, with auxiliary estimates being mostly beneficial to MPNN. Furthermore, there are a few occurrences where some augmentations, which did help individually, were better removed from fully augmented models. This suggests that 1) the weighting of the auxiliary tasks may need to be optimised, and 2) some augmentations may interact and result in a more complex behaviour. Total model complexity may also need to be accounted for to explain these behaviours. Future work will further explore the cause for these varying behaviours and avenues to better balance the contributions of auxiliary estimations for example based on \citep{Long2017,Cipolla2018} instead of the fixed $\beta=0.3$. For the rest of the paper, since the combined four augmentations still outperform base models by a significant margin, we use this simple configuration to assess the properties of fully augmented GNNs.

\subsection{Evaluation of the fully augmented models on OoE molecule and crystal data}
\label{subsec:comparison_MPNN}

\begin{figure}
    \centering
    \includegraphics[width=0.8\linewidth]{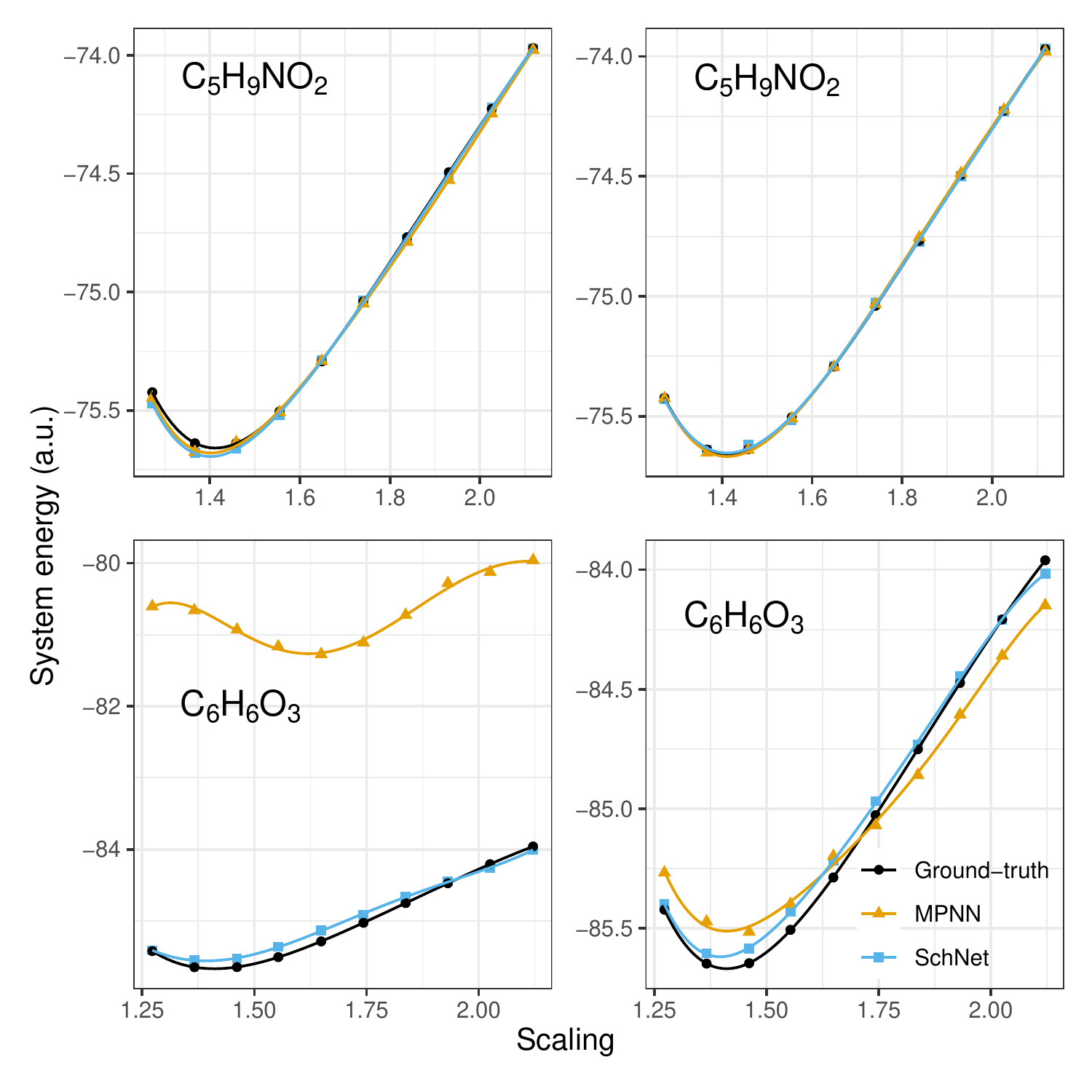}
    \caption{Energy estimations at different scalings of two examples of molecules for base (left) and augmented (right) GNNs.}
    \label{fig:curves_molecules}
\end{figure}

\begin{figure}
    \centering
    \vspace{10pt}
    \includegraphics[width=0.8\linewidth]{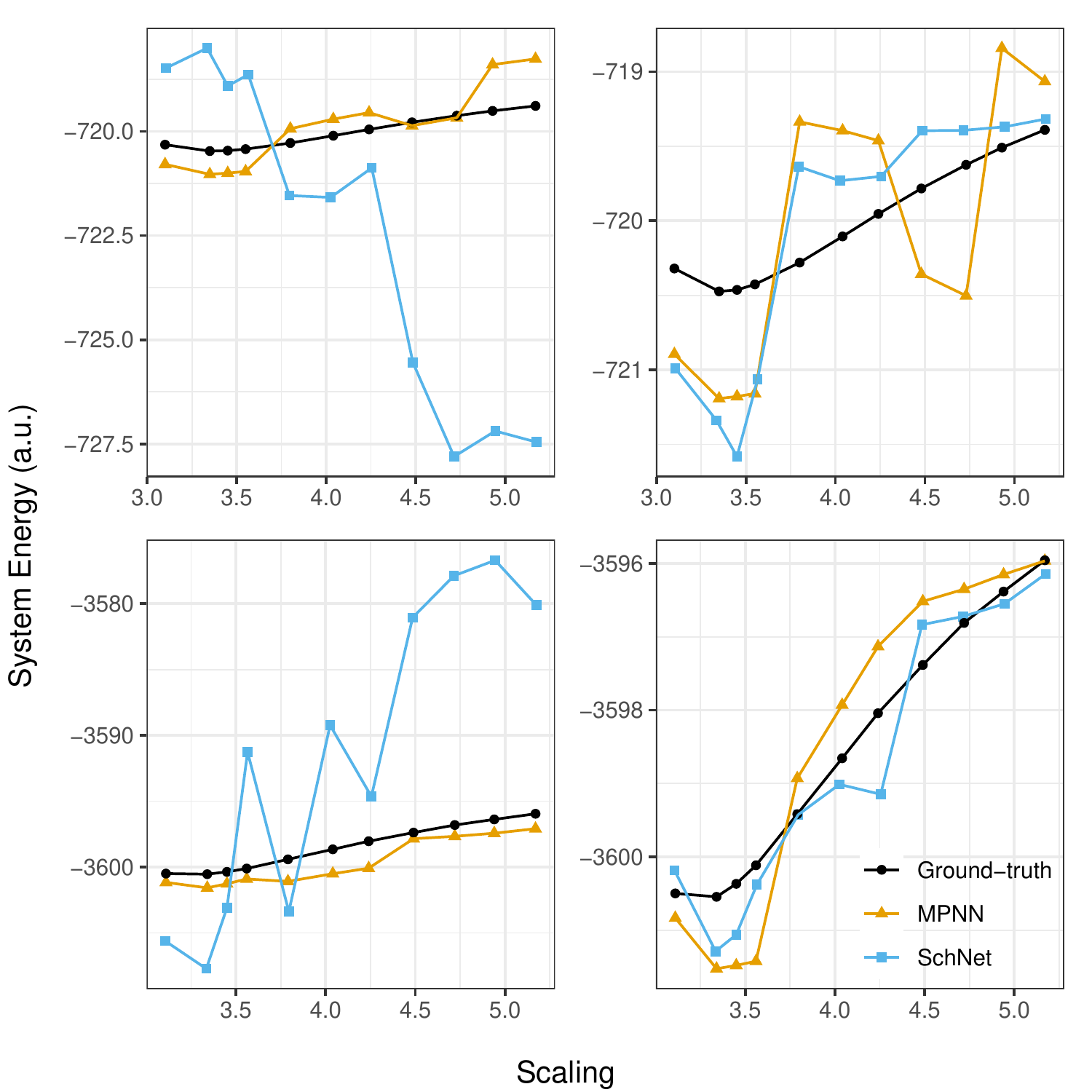}
    \caption{Example energy estimations at different scalings for a small (15 atoms, top) and large (75 atoms, bottom) crystalline system. Left: base models, right: augmented models}
    \label{fig:curves_crystals}
\end{figure}

We combine all our domain integration methods (i.e. the best performing method for $r$-specialised interactions and all auxiliary estimates) into Augmented-MPNN and Augmented-SchNet.
For $r$-specialised interactions, we retain Eq.~\ref{eq:specialised_update} impl. 2 for MPNN and Eq.~\ref{eq:weighted_sum_messages} with scalar $\lambda_r$ for SchNet. Given the complexity discussion of the previous section, we will also present some results using the simpler (and still reasonably well performing) Eq.~\ref{eq:weighted_sum_messages} with scalar $\lambda_r$ for MPNN.
Although $r$-specialised message production and $r$-specialised node update are not mutually exclusive, their combination will be explored in future work.
For CG, we consider system sizes of 15 to 75 atoms instead of the available 114 due to memory restrictions. Results are reported in Table~\ref{tab:datasets} and with more details and metrics in the sup. materials.
We further illustrate the performance of the fully augmented GNNs by plotting their energy estimates for all scaled versions of randomly picked systems of E-QM9 in Fig.~\ref{fig:curves_molecules}, and of CG in Fig.~\ref{fig:curves_crystals}.

We first note that it is beneficial to combine $r$-specialised interactions and auxiliary estimations, with AE and/or DSG results being further improved in Table~\ref{tab:datasets} as compared to individual augmentations in Table~\ref{tab:branches}. The fact that some auxiliary estimations are chosen to support $r$-specialised interactions may be a reason for these further improvements.

Both GNN types benefit from our domain knowledge integration, with overall more accurate energy estimates and better performance at finding stable geometries (DSG). The latter point is also illustrated in the example energy plots of Figs.~\ref{fig:curves_molecules} and \ref{fig:curves_crystals} where the augmented GNNs produce curves with a correct minimum while base GNNs sometimes had physically-unrealistic curves with incorrect minimum.
The improvement is particularly strong on CG, hinting that our augmented models can generalise better to new geometries, that are a particularity of this dataset.
Consequently, on Periodic Crystals, which contains only two mono-atomic crystals (with various scalings), we expect to see further improvements in future works by considering structures outside of the fcc lattice and a wider variety of atom types.

\subsection{Interpretation of the nodes' hidden states}
\label{subsec:visualisation}

\begin{figure}[h]
    \centering
    \includegraphics[width=0.9\linewidth]{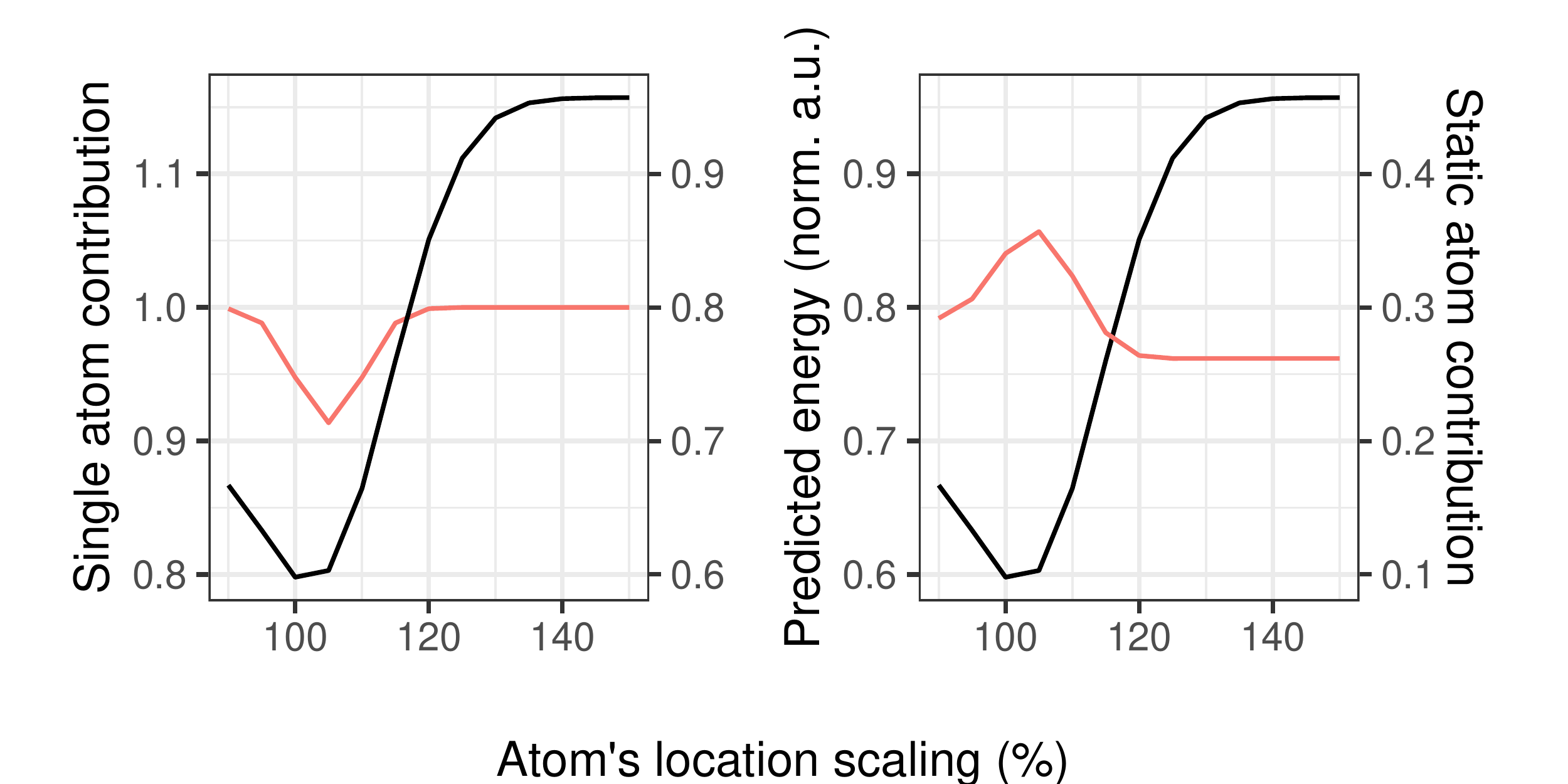}
    \caption{Contribution of a moving atom (left red curve) and mean contribution of static atoms (right red curve) toward the energy estimate of a crystal (black curves) by Augmented-SchNet}
    \label{fig:visualisation}
\end{figure}

We examine the node states through visualising the contribution of each node to the final estimate. In MPNN, perceptrons combine the states of a node at all timesteps. In SchNet, the last atom-wise layer has output size of 1. In both networks, these reduced values are summed across nodes to obtain the energy estimate. After normalising across all nodes, they may be considered as the \emph{atoms' contributions} to the energy estimate.

After training the GNNs on UCG, we consider an atom that is newly added to a (stable) crystal seed of 14 atoms. When scaling the distance of this single atom to the rest of the crystal, we examine its contribution and that of the rest of the (static) atoms.
This scenario differs from the training scenario of whole system scaling, but Augm.-SchNet in Fig.~\ref{fig:visualisation} still produces plausible energy estimates with correct minimum (although some offsets in energy sometimes happen and will be investigated in future work).

As shown in the sup. materials, base GNNs were not successful, including SchNet although this `freely' moving atom setup is closer to its original experimental setup \citep{Schutt2017Schnet:Interactions}, possibly due to not being able to learn on scaled systems and many sizes at once. Thus, our augmentations increased the generalisation ability of SchNet through a better capture of the atomic interactions.

\begin{figure*}[h]
    \centering
    \includegraphics[width=0.9\textwidth]{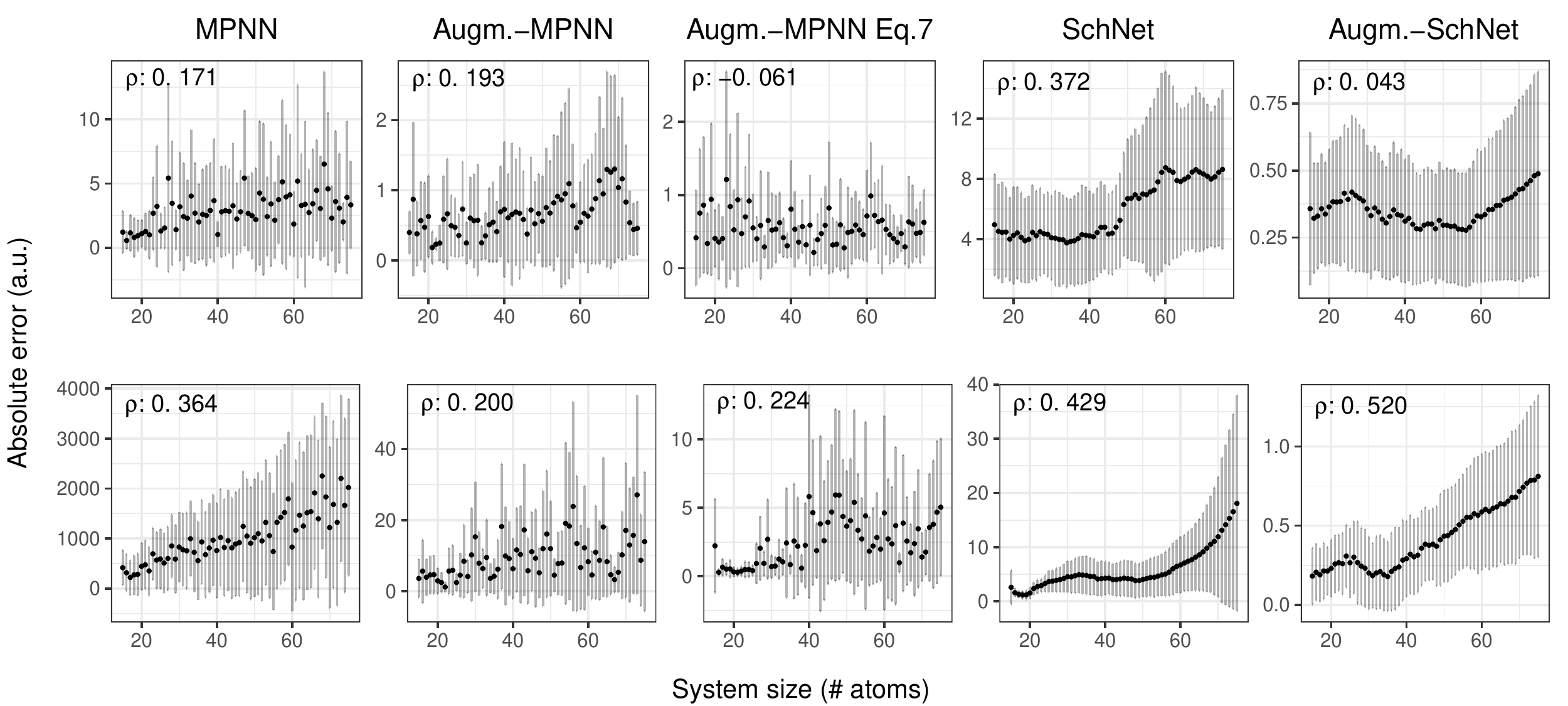}
    \caption{Impact of system size on accuracy when trained on systems of up to 75 (top) and 25 (bottom) atoms from UCG.}
    \label{fig:outside}
\end{figure*}

Augm.-MPNN could not obtain a plausible energy curve with correctly located minimum either. We tested both the best performing specialised interaction method of Eq.~\ref{eq:specialised_update}, that introduces a higher model complexity, and the simpler specialised interaction method of Eq.~\ref{eq:weighted_sum_messages} with scalar $\lambda_r$, used in Augm.-SchNet. Neither methods succeeded in handling the single moving atom, which indicates that the lower generalisation to new scenarios does not come from the higher complexity introduced by Eq.~\ref{eq:specialised_update}, but rather would be inherent to the MPNN model, and could not be overcome by our augmentations as in SchNet.

In Augm.-SchNet, the moving atom's contribution (left red curve) is minimal at its stable location (i.e. minimum of energy, black curve in Fig.~\ref{fig:visualisation}). At the same time, static atoms (right red curve) contribute maximally, in line with the crystal being at its minimum energy driven by its regular structure. As the atom is moved closer or pulled away, its contribution increases simultaneously with energy. At the same time, the relative contributions of static atoms decrease to give way to the perturbation of the moving atom. This strongly suggests that the GNN learnt to pay attention to the location of individual atoms, although it was trained on systems that are isometrically scaled.

The same experiment, when performed after training on SCG, is not as successful, with Augm.-SchNet producing an implausible energy curve with no minimum. This indicates that the different geometries provided by the 20 varied locations of atoms for each crystal size were not enough to learn the importance of fine location of individual atoms. When training on UCG, the (global) scaling was complementary to the varied occupations of atom sites in learning the importance of precise location of individual atoms.

It is worth noting that the same training did not allow original SchNet to learn this principle and to produce plausible energy curves. Thus, it is the combination of our augmentations and of a sufficiently diverse training data that achieved this result.

\subsection{Generalisation to larger graphs}
\label{subsec:generalisation_test}

Fig.~\ref{fig:outside} presents energy AE on UCG as a function of system size for base and augmented GNNs when trained on all system sizes (up to 75 atoms) or on small systems only (up to 25 atoms), keeping the testing set equal.
For Augm.-SchNet, the better accuracy from the augmentations maintains the AE in the best achieved range when training on small systems only.
For Augm.-MPNN, when training on small systems only, the range of AE is also improved by a factor $\sim$100 compared to base MPNN.
Furthermore, accuracy correlates with system size only marginally stronger than when training on all sizes (Pearson coef. 0.200 against 0.193). This is an improvement from base MPNN, where Pearson correlation increases from 0.171 to 0.364. Therefore, our augmentations allow MPNN for learning of some basic principles about the atomic interactions that are applicable to larger, unfamiliar systems.
When using the specialised interaction method of Eq.~\ref{eq:weighted_sum_messages} with scalar $\lambda_r$, Pearson correlation is similar at 0.224, and the range of AE is improved by a factor $\sim$400. We explain this better AE by the different complexities of Eqs.~\ref{eq:specialised_update} and \ref{eq:weighted_sum_messages}, which make the model more or less robust to smaller training sets.

\subsection{Generalisation to smaller training sets}
\label{subsec:generalisation_test_2}

\begin{figure}[h]
  \centering
  \includegraphics[width=0.8\linewidth]{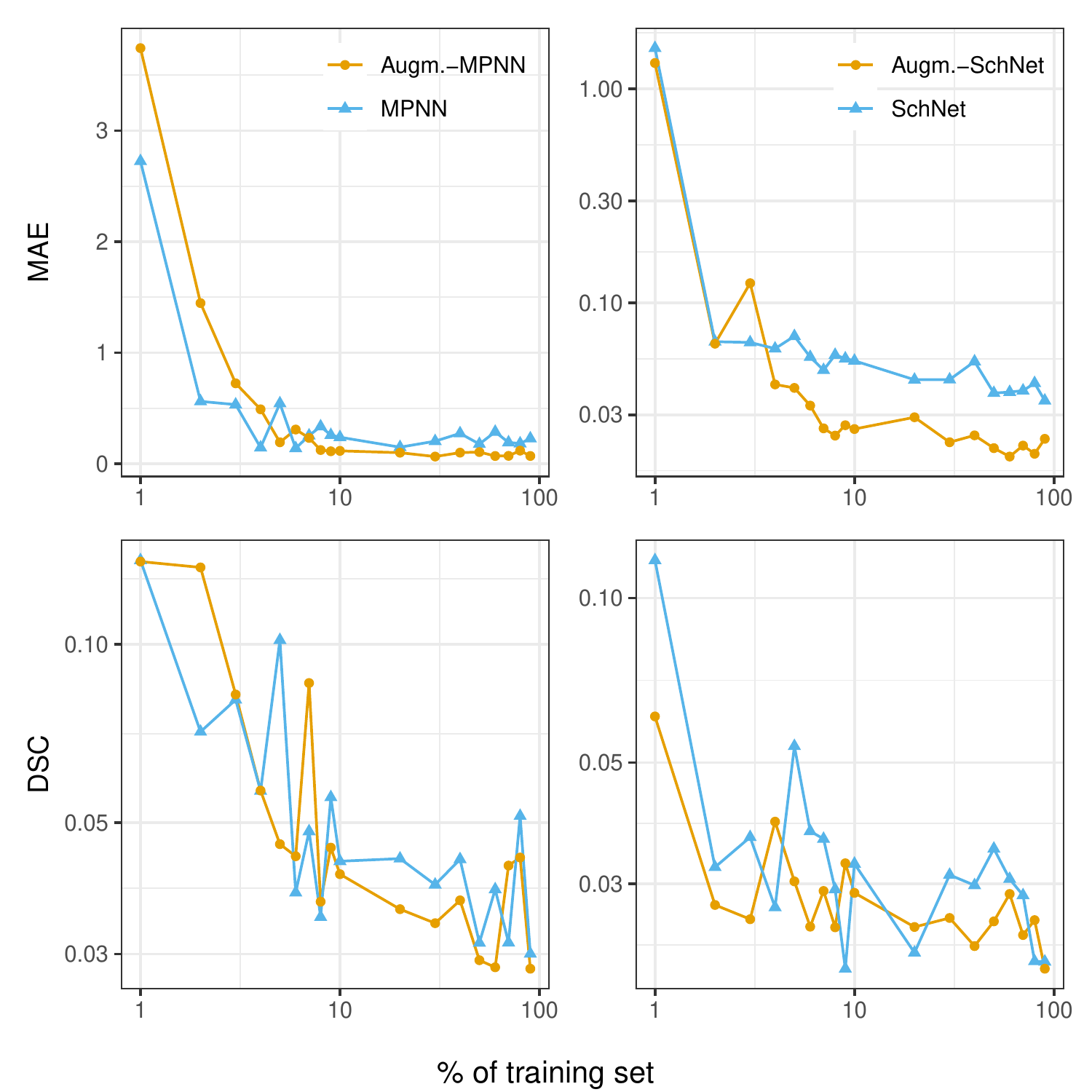}
  \caption{Robustness of base and augmented GNNs to smaller training sets (i.e. subsets of molecules) for E-QM9. Top: mean AE, bottom: mean DSG. Left: MPNN, right: SchNet. Blue: base, orange: augmented.}
  \label{fig:reduced_mol_EQM9}
\end{figure}

\begin{figure}[h]
  \centering
  \includegraphics[width=0.8\linewidth]{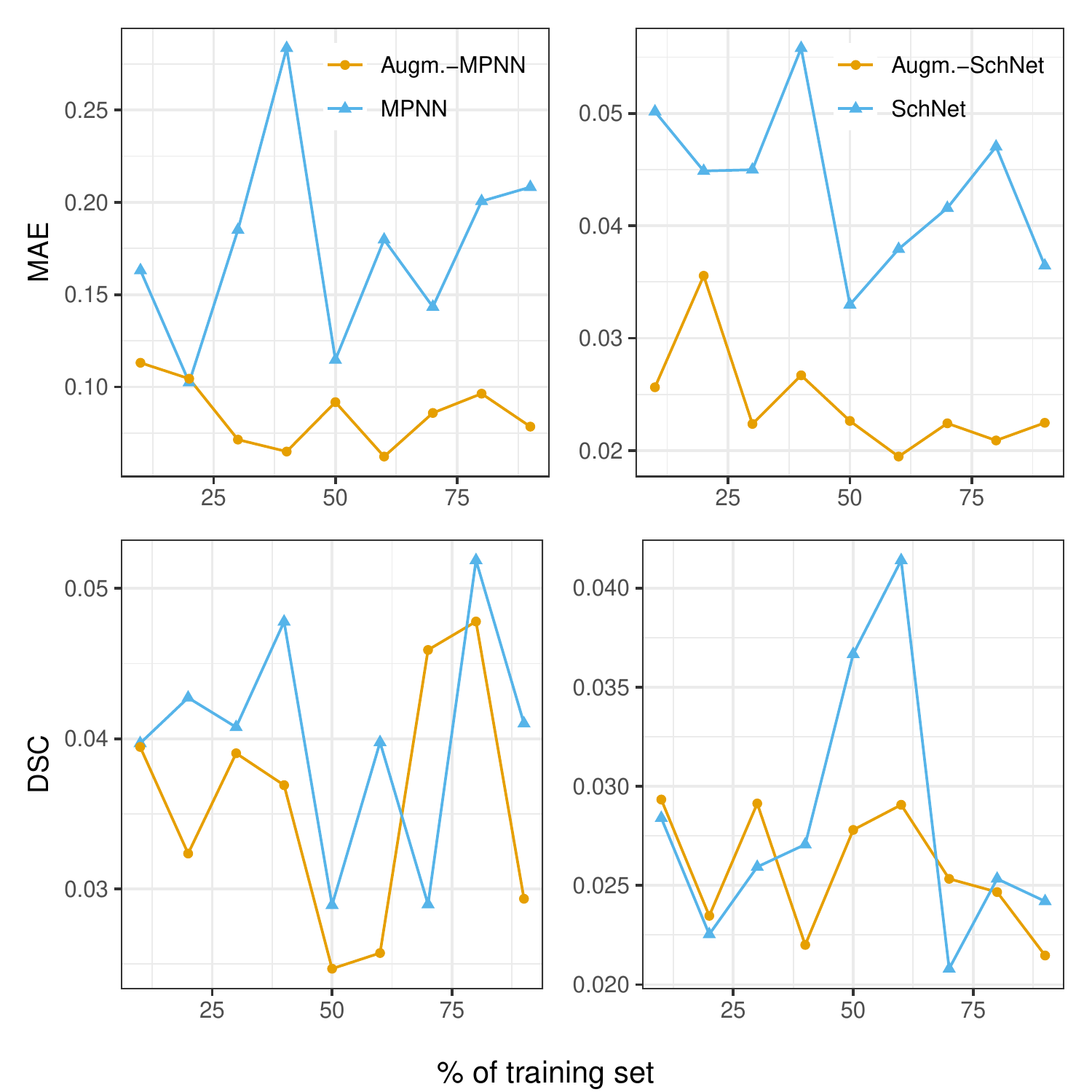}
  \caption{Robustness of base and augmented GNNs to smaller training sets (i.e. subsets of scalings) for E-QM9. Top: mean AE, bottom: mean DSG. Left: MPNN, right: SchNet. Blue: base, orange: augmented.}
  \label{fig:reduced_scal_EQM9}
\end{figure}

\begin{figure}[h]
  \centering
  \includegraphics[width=.8\linewidth]{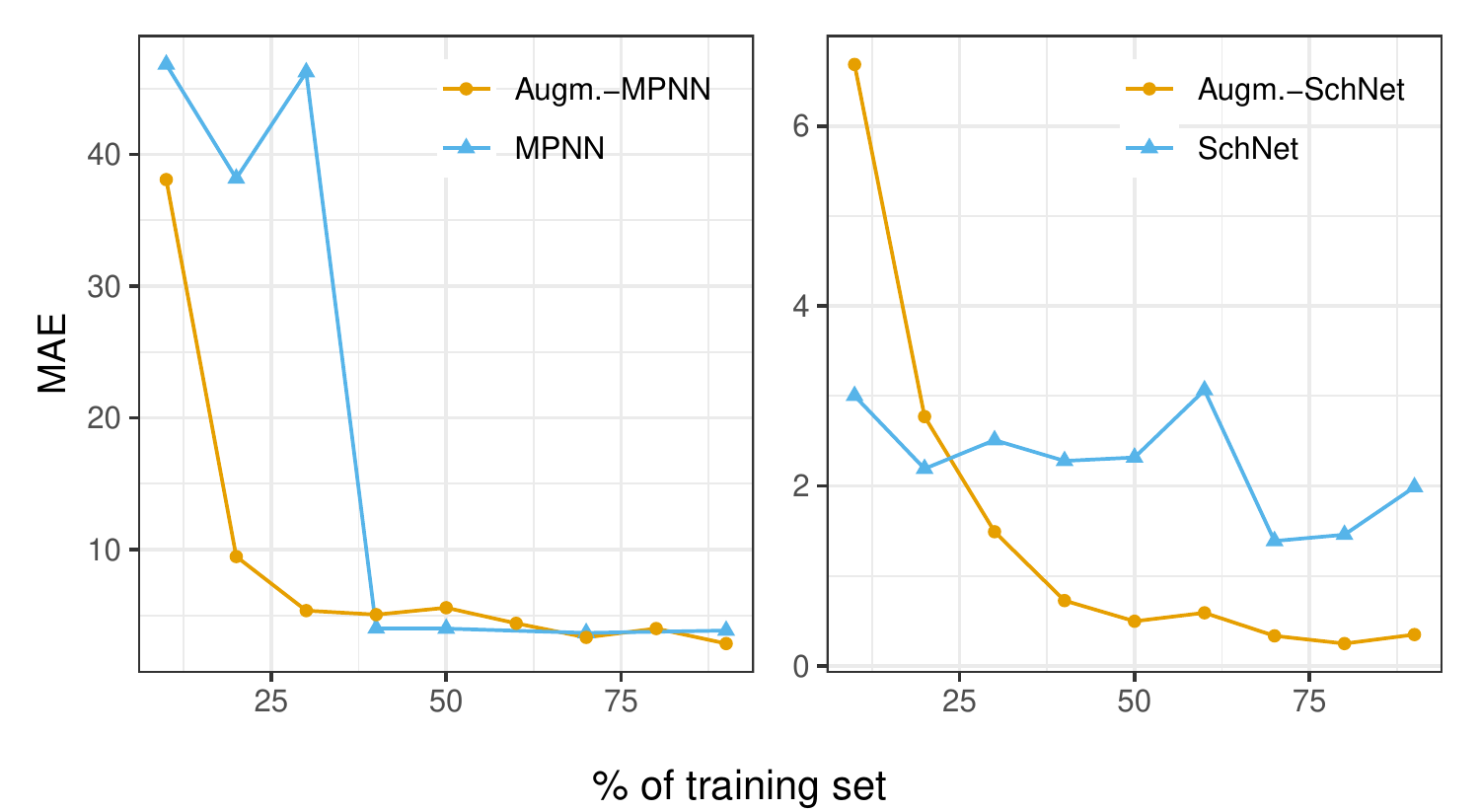}
  \caption{Robustness of base and augmented GNNs to smaller training sets for SCG. Left: MPNN, right: SchNet. Blue: base, orange: augmented.}
  \label{fig:reduced_SCG}
\end{figure}

We test whether the domain knowledge integration brings a higher robustness to small training sets through a better generalisation, by reducing progressively the size of the training set (keeping the testing set equal).
Figs.~\ref{fig:reduced_mol_EQM9}, \ref{fig:reduced_scal_EQM9}, and \ref{fig:reduced_SCG} present results respectively on E-QM9 when decreasing the variety of molecules and when decreasing the variety of scalings, and on SCG (full details of the experimental setup are in the sup. materials).

When considering extremely small training sets, there are a few cases where the augmented GNNs obtain worse results than base GNNs. This might be explained by the increased model complexity due to additional learnt parameters and auxiliary tasks, requiring a minimal amount of data to train effectively.

More generally, on E-QM9, while augmented GNNs tend to outperform their base counterparts, they are not more robust to decreasing the number of molecules to be trained on as performances decrease with similar rates (Fig.~\ref{fig:reduced_mol_EQM9}). Thus, the augmentations could not compensate for the loss of composition diversity.

However, both augmented GNNs are quite stable to varying numbers of scalings, and more so than their base counterparts (Fig.~\ref{fig:reduced_scal_EQM9}). This indicates that they are better able to account for the effect of varying geometries.

The most dramatic improvement is on SCG, where both augmented GNNs see a more stable, then a later rise in mean AE when decreasing the dataset size and the associated diversity in crystal geometries (Fig.~\ref{fig:reduced_SCG}).
We conclude that the augmentations provided insights on geometry and scaling that did not need to be learnt from data, hence reducing the need for exhaustively covering these aspects in training samples.

\section{Conclusion}
\label{sec:conclusion}

We integrate domain knowledge into GNNs to improve accuracy and generalisation, with two proposed strategies: (1) specialised information flow within the GNN to better account for relation types between nodes, and (2) further relating the learnt representations to the studied phenomenon through MTL. We explore the different means for specialising information flow by acting on either message production or node update, and we provide general formulations that are adapted to different architectures. We demonstrate them on two architectures: MPNN and SchNet.

Our proposed domain knowledge integration is tested on a quantum chemistry case study where potential energy of chemical systems (molecules or crystals) is estimated as a function of geometry.
For this application, the elements of domain knowledge that we use are specifically related to atomic interaction and underlying physics of potential energy.

The added model complexity varies widely with the augmentation methods, and a too high complexity for a given GNN and application may be detrimental. However, in many cases, the added complexity proved helpful in learning complex and more general concepts.
It allowed the GNNs to better learn principles of atomic interactions, with improved handling of unseen geometries in graphs, including unseen sizes of graphs and unseen perturbations of their nodes' locations, and the ability to train on smaller datasets.

Our domain knowledge integration methods add to the toolbox of GNN augmentation strategies.
They are generally applicable to non-physics domains, such as economics \citep{Panford2020}, traffic prediction \citep{Diehl2019}, or predicting relationships for E-commerce recommendation \citep{Zhang2019}, where problems are naturally represented as graphs. These new applications of our augmentations would require an expert-based identification of relations and auxiliary tasks of interest, following the example of our case study.

From the viewpoint of our case study, our augmentations may enhance current and future state-of-the-art (SoTA) in estimating potential energies and stable geometries of chemical systems. Indeed, current SoTA are based on the two seminal GNNs used in our case study, and we expect similar results may be obtained from augmenting their more recent variations. Furthermore, our results on two very different GNN architectures suggest that similar improvements may also be obtained on other GNN types, including future GNN-based SoTAs.
However, it is well-known that DNNs cannot offer guarantees on results and suffer from well documented problems such as adversarial examples. Therefore, although our methods improve their accuracy and robustness, numerical simulations remain necessary for a final verification of an optimal geometry found through a DNN's predictions.



\bibliographystyle{elsarticle-harv} 
\bibliography{cas-refs}





\end{document}